%% file: main.tex
\documentclass[letterpaper,10pt,conference]{IEEEtran}

\IEEEoverridecommandlockouts

\usepackage{combelow}

\usepackage{glossaries}
\newacronym{fmcw}{FMCW}{Frequency-Modulated Continuous-Wave}
\newacronym{nn}{NN}{neural network}
\newacronym{fft}{FFT}{Fast Fourier Transform}
\newacronym{cnn}{CNN}{Convolutional Neural Network}

\usepackage{cite}

\usepackage{graphicx}

\usepackage{paralist}

\usepackage[labelformat=simple]{subcaption}

\usepackage[hidelinks]{hyperref}
\newcommand\rurl[1]{%
  \texttt{\href{http://#1}{\nolinkurl{#1}}}%
}

\usepackage{cleveref}
\crefname{table}{Tab.}{Tabs.}
\crefname{figure}{Fig.}{Figs.}
\crefname{section}{Sec.}{Secs.}
\crefname{equation}{Eq.}{Eqs.}

\usepackage[binary-units]{siunitx}
\usepackage{pgfplotstable}
\usepackage{tikz}

\usepackage{mathabx}

\usepackage{xcolor}

\begin{document}

%------------------------------------------------------------------
\title{
\large \bf Unsupervised Place Recognition with Deep Embedding Learning over Radar Videos
}
\author{
Matthew Gadd$^{\dagger}$, Daniele De Martini$^{\dagger}$, and Paul Newman\\
Oxford Robotics Institute, Dept. Engineering Science, University of Oxford, UK.\\\texttt{\{mattgadd,daniele,pnewman\}@robots.ox.ac.uk}
\thanks{
$^{\dagger}$ Equal contribution.
This work was supported by the Assuring Autonomy International Programme, a partnership between Lloyd’s Register Foundation and the University of York.
}
}
\maketitle
%------------------------------------------------------------------

%------------------------------------------------------------------
\begin{abstract}
We learn, in an unsupervised way, an embedding from sequences of radar images that is suitable for solving place recognition problem using complex radar data.
We experiment on \SI{280}{\kilo\metre} of data and show performance exceeding state-of-the-art supervised approaches, localising correctly \SI{98.38}{\percent} of the time when using just the nearest database candidate.
\end{abstract}
\begin{IEEEkeywords}
Radar, Place Recognition, Deep Learning, Unsupervised Learning, Autonomous Vehicles
\end{IEEEkeywords}

%------------------------------------------------------------------
\section{Introduction}%
\label{sec:introduction}
%------------------------------------------------------------------

For autonomous vehicles to drive safely at speed, in inclement weather, or in wide-open spaces, very robust sensing is required.
Thus, the interest in scanning \acrshort{fmcw} radar for place recognition and localisation.
As more datasets featuring this modality become available, unsupervised techniques will allow us to learn from copious unlabelled measurements.
This paper presents initial results in that vein.

%------------------------------------------------------------------
\section{Related Work}%
\label{sec:related}
%------------------------------------------------------------------

Kim \textit{et al}~\cite{kim2020mulran} show that radar can outperform LiDAR in place recognition using a non-learned rotationally-invariant ring-key descriptor.
S{\u{a}}ftescu \textit{et al}~\cite{saftescu2020kidnapped} present the first supervised deep approach, adapting \acrshortpl{cnn} for equivariance and invariance to azimuthal perturbations.
Barnes and Posner~\cite{barnes2020under} learn keypoints which are useful for motion estimation and localisation simultaneously.
Gadd \textit{et al}~\cite{gadd2020lookaroundyou} leverage rotationally-invariant representations in a sequence-based localisation system.
De Martini \textit{et al}~\cite{demartini2020kradar} fine-tune the results from the embedding space with pointcloud registration techniques.
Wang \textit{et al} learn metric localisation directly along with self-attention~\cite{wang2021radarloc}.

%------------------------------------------------------------------
\section{Method}%
\label{sec:method}
%------------------------------------------------------------------

\begin{figure}
\centering
\includegraphics[width=0.98\columnwidth]{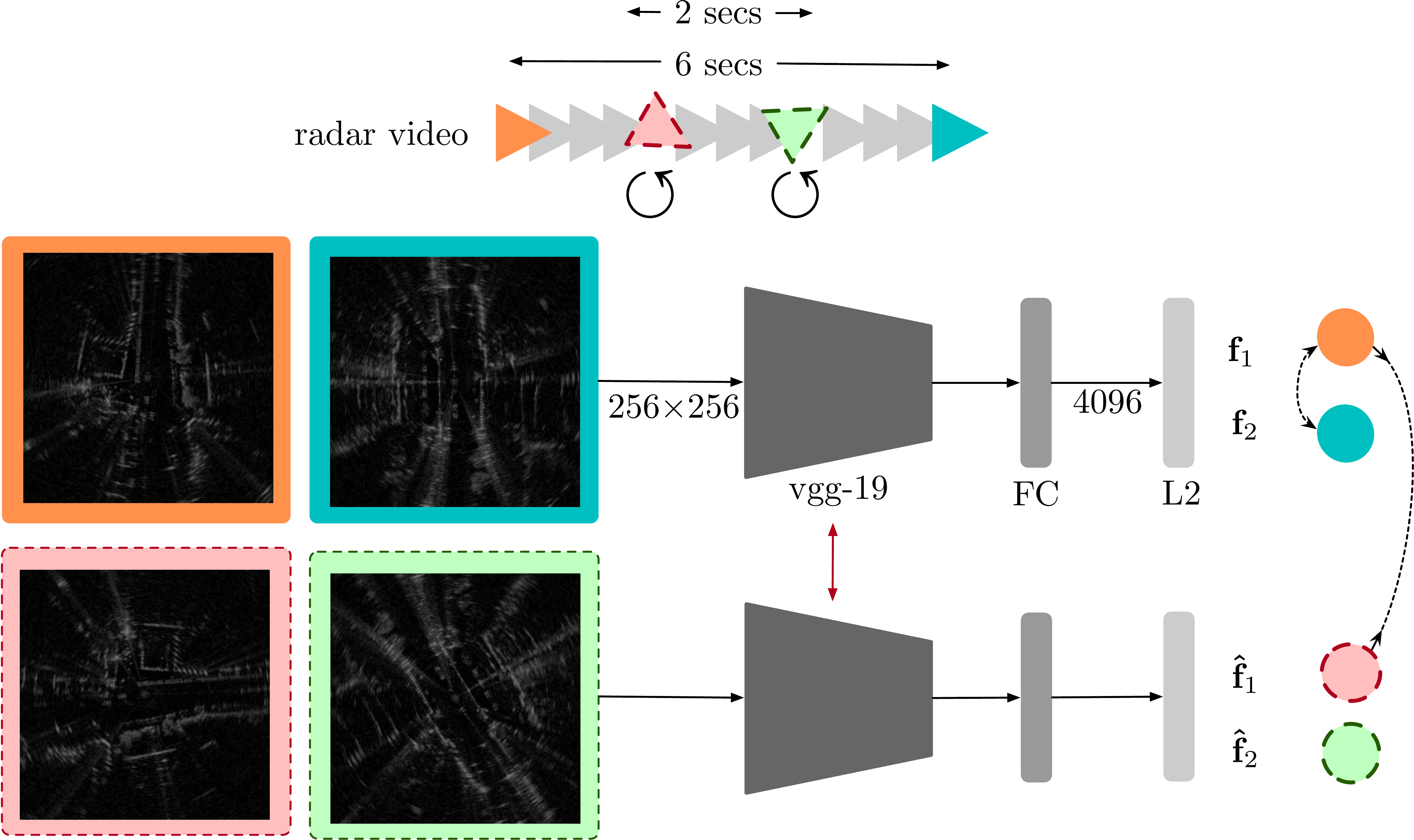}
\caption{
Overview of our network architecture and the training procedure more generally -- described in~\cref{sec:method}.
}
\label{fig:system}
\vspace{-0.5cm}
\end{figure}

Our method is illustrated in~\cref{fig:system}.
We adapt~\cite{ye2019unsupervised}, which ensures that features of distinct instances (originally, camera images) are separated (right, arrow line between orange and blue) while features of an augmented instance are invariant (right, back-arrow line between orange and red).
Our \emph{contributions} are the sampling \emph{as well as} augmentation strategies (top) most apposite to constrain the learning of these features for this sensor modality and the place recognition task.
Here, in constructing batches of instances and their augmentations (left), we:
\begin{enumerate}
\item\label{meth:inst} sample an instance randomly (orange),
\item\label{meth:nearby} pin the pre-augmented instance at \SI{2}{\sec} later (red),
\item\label{meth:rotate} augment either~\ref{meth:inst} or~\ref{meth:nearby} (c.f.~\cref{sec:experiments}, \texttt{vR} and \texttt{vT}) by spinning (red, $\circlearrowleft$),
\item\label{meth:tn} pin \emph{another} instance at \SI{6}{\sec} later (blue),
\item\label{meth:augn} spin \emph{its} augmentation from \num{2} seconds before (green, $\circlearrowleft$).
\end{enumerate}

\cref{sec:experiments,sec:results} prove the effectiveness of each step.
Briefly, however, consider the following rationale.
In step~\ref{meth:nearby} we ensure that real data is presented to the network, rather than synthetic augmentations which would anyway be prohibitively difficult to simulate for radar sensor artefacts.
This step also presents the network with some translational invariance, along the route driven.
In step~\ref{meth:rotate} we ensure rotational invariance.
In step~\ref{meth:tn} we are mitigating the (already quite small but non-zero) chance that batches may simultaneously hold instances which are in fact nearby, although randomly sampled.
The combination of steps~\ref{meth:tn} and~\ref{meth:augn} prepare this ``true negative'' as just another perturbed radar scan as per steps~\ref{meth:inst},~\ref{meth:nearby}, and~\ref{meth:rotate}, but which usefully constrains the learning in this way.

For completeness, we note that we use the \texttt{VGG-19} feature extractor~\cite{simonyan2014very}.
We train each variant for 10 epochs, using a learning rate of $3\mathrm{e}{-4}$, batch size of 12.
Images are presented to the network in Cartesian format with dimensions $256{\times}256$ where the side-length of each pixel represents \SI{0.5}{\metre}.

%------------------------------------------------------------------
\section{Experimental Setup}%
\label{sec:experiments}
%------------------------------------------------------------------

We evaluate our method in training and testing across urban data collected in the \textit{Oxford Radar RobotCar Dataset}~\cite{barnes2020rrcd}.
Train-test splits are as per~\cite{saftescu2020kidnapped}, testing on hidden data featuring backwards traversals, vegetation, and ambiguous urban canyons.
We do, however, train on most data available from~\cite{barnes2020under} (barring anything from the test split) -- totalling \num{30} forays, or about \SI{280}{\kilo\metre} of driving.
This is a similar quantity of training data as reported in~\cite{barnes2020under}.

We examine the performance gain available from:
\begin{itemize}
\item \texttt{vR} or ``spin augmenting'': steps~\ref{meth:inst} and \ref{meth:rotate} (spinning the instance from step~\ref{meth:inst}),
\item \texttt{vT} or ``video sampling'': steps~\ref{meth:inst} and \ref{meth:nearby} (no spinning),
\item \texttt{vTR} or ``video sampling and spin augmenting'': steps~\ref{meth:inst}, \ref{meth:nearby}, and \ref{meth:rotate}, and finally
\item \texttt{vTR2}: steps~\ref{meth:inst}, \ref{meth:nearby}, \ref{meth:rotate}, \ref{meth:tn}, and \ref{meth:augn}, which we refer to as ``video sampling and spin augmenting with batch pairing''.
\end{itemize}

Note the baseline of~\cite{ye2019unsupervised}, entails augmenting \emph{only} the instance itself, closest in spirit to \texttt{vR}.

For performance assessment, we employ both \texttt{Recall@P} -- where predicted positives are within in a varying embedding distance threshold of the query -- and \texttt{Recall@N} -- where a query is considered correctly localised if \emph{at least one} database candidate is close in space as measured by GPS/INS.
Note that the latter is a more lenient measure of performance and that we employ no pose refinement to boost results, as in~\cite{kim2020mulran,demartini2020kradar}.
We specifically impose for the first metric (\cref{fig:ball-search}) boundaries of \SI{25}{\metre} and \SI{50}{\metre}, inside of which predicted matches are considered true positives, outside of which false positives, as per~\cite{saftescu2020kidnapped}.
The second metric (\cref{fig:db-cand}) supports only a single boundary, which we set to \SI{25}{\metre}, the more difficult of the two.

\begin{figure}
\centering
\begin{subfigure}{0.4\columnwidth}
\includegraphics[width=\textwidth]{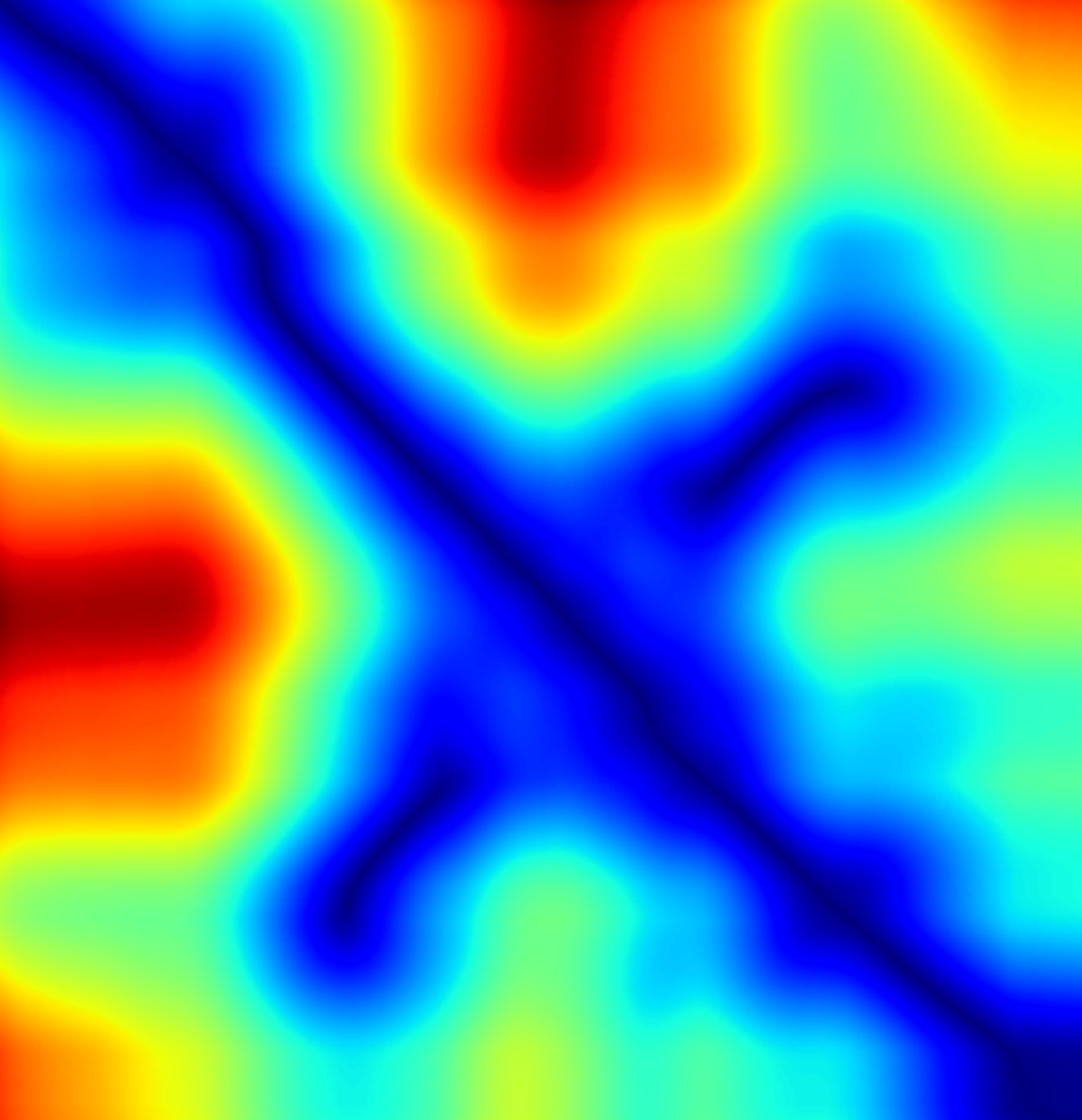}
\caption{}
\label{fig:gt}
\end{subfigure}
\begin{subfigure}{0.4\columnwidth}
\includegraphics[width=\textwidth]{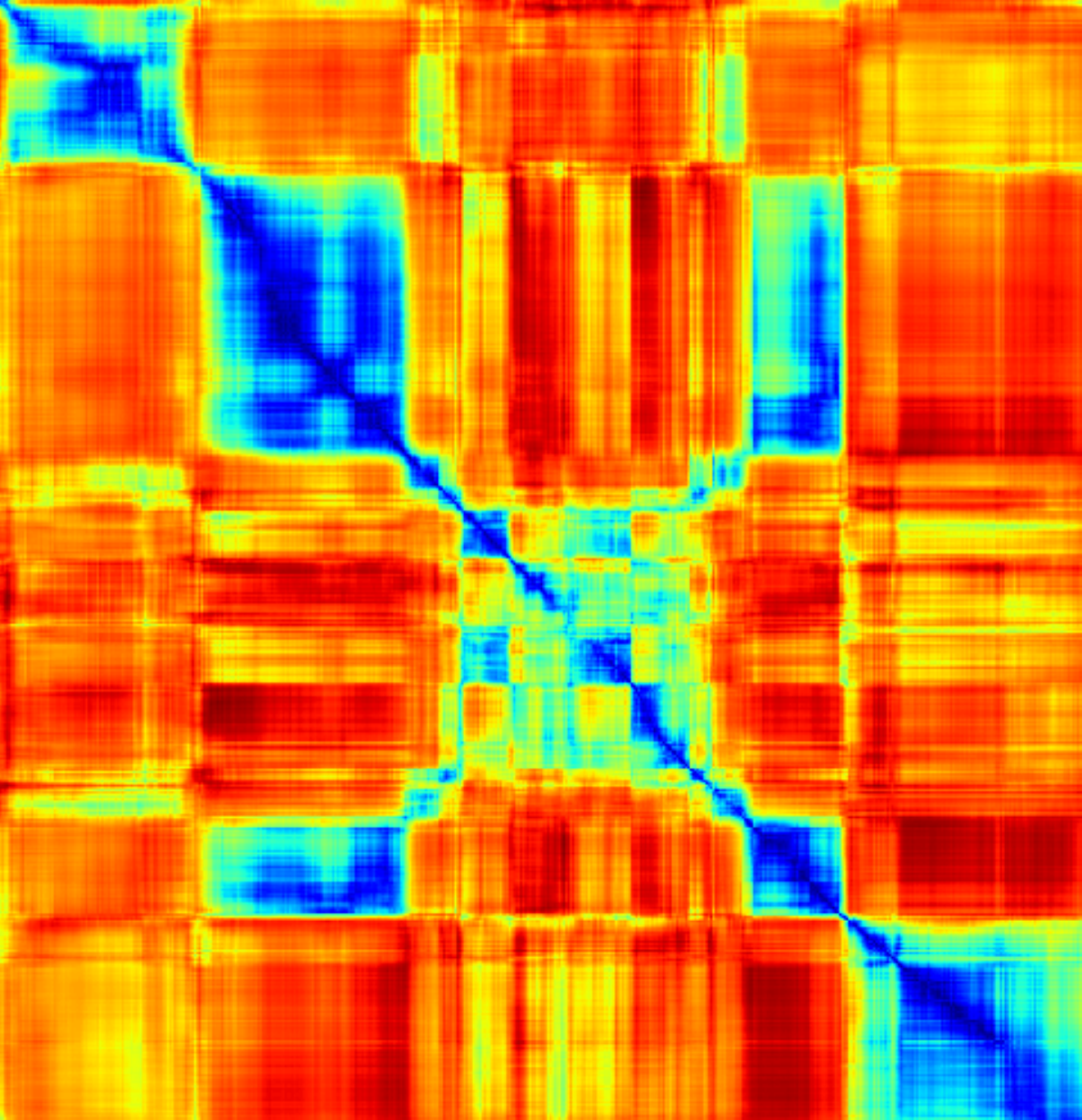}
\caption{}
\label{fig:vTR2_dist}
\end{subfigure}
\begin{subfigure}{0.4\columnwidth}
\includegraphics[width=\textwidth]{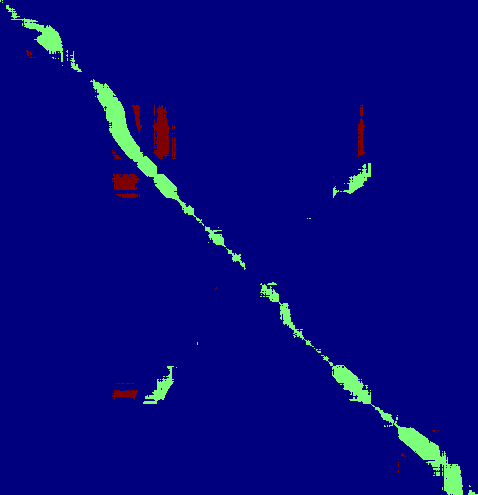}
\caption{}
\label{fig:bsnn}
\end{subfigure}
\begin{subfigure}{0.4\columnwidth}
\includegraphics[width=\textwidth]{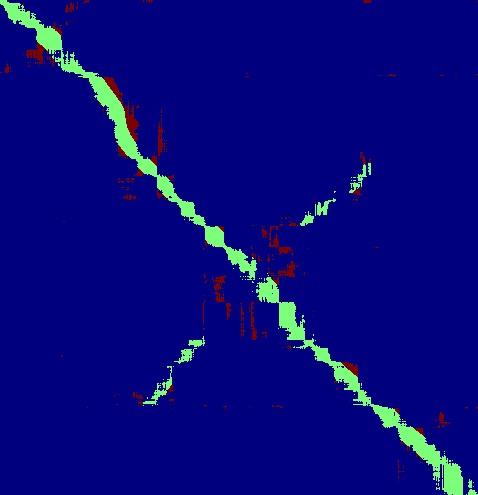}
\caption{}
\label{fig:rnnn}
\end{subfigure}
\caption{
\subref{fig:gt} Ground truth matrix,
\subref{fig:vTR2_dist} Distance matrix for embeddings learned by our top-performing model, \texttt{vTR2}, 
\subref{fig:bsnn} match matrix for \texttt{Recall@P=80\%},
\subref{fig:rnnn} match matrix for \texttt{Recall@N=25}.
In \subref{fig:bsnn} and \subref{fig:rnnn}, true positives are shown green while false positives are shown red.
}
\label{fig:mats}
\vspace{-.4cm}
\end{figure}

%------------------------------------------------------------------
\section{Results}%
\label{sec:results}
%------------------------------------------------------------------

\input{plots/ball-search}
\input{plots/db-cand}

\cref{fig:ball-search} shows the \texttt{Recall@P} metrics.
We achieve from \SI{11.72}{\percent} up to as much as \SI{16.34}{\percent} recall for precisions of \SIrange{99}{95}{\percent}.
This is compared to \SIrange{4.26}{7.08}{\percent} for the baseline, \texttt{vR}.
This corresponds to $F_1$, $F_2$, and $F_{0.5}$ scores of \num{0.60}, \num{0.55}, and \num{0.63}, respectively.
Here, we have outperformed the supervised~\cite{saftescu2020kidnapped}, where on this same test split the corresponding scores were reported as \num{0.53}, \num{0.48}, and \num{0.60}.

\cref{fig:db-cand} shows the \texttt{Recall@N} metrics.
Here, we notice when only using \num{1} database candidate, we correctly localise \SI{98.38}{\percent} of the time (\texttt{vTR2}) as compared to \SI{73.13}{\percent} for the baseline, \texttt{vR}.
This is also superior to the results reported in~\cite{barnes2020under}, approximately \SI{97}{\percent}.

Note the fidelity of the embedding space for \texttt{vTR2},~\cref{fig:vTR2_dist}, and the ground truth matrix, ~\cref{fig:gt}.
Differently to \cref{fig:ball-search,fig:db-cand}, \cref{fig:bsnn} shows the true and false positives for \texttt{Recall@P=80\%}.
\cref{fig:rnnn} shows such for \texttt{Recall@N=25}, similarly to~\cite{kim2020mulran,barnes2020rrcd}.
In both, we consider it reasonable to expect a downstream process to disambiguate \emph{some} false matches in order to retain various types of loop closure. 
We see that rotational invariance is well understood by the network trained as we propose.

%------------------------------------------------------------------
\section{Conclusion}%
\label{sec:conclusion}
%------------------------------------------------------------------

We have presented our initial findings for applying unsupervised learning to radar place recognition.
In doing so, we outperform two previously published methods which are candidates for the current state-of-the-art, showing the great promise of unsupervised techniques in this area.

%------------------------------------------------------------------
\bibliographystyle{IEEEtran}
\bibliography{biblio}
%------------------------------------------------------------------

\end{document}

%% file: plots/ball-search.tex
\pgfplotsset{
select row/.style={
x filter/.code={\ifnum\coordindex=#1\else\def\pgfmathresult{}\fi}
}
}

\pgfplotstableread[col sep=comma]{
metric,vT,vR,vTR,vTR2
95,7.081667618466944,0.5477389543608965,12.400706089962094,16.349099215947515
98,4.262499350998065,0.5477389543608965,9.041586625777262,13.208036965821046
99,4.262499350998065,0.4880328124163437,7.533357561873561,11.728363013282133
}\datatable

\begin{figure}
\centering
\begin{tikzpicture}
\begin{axis}[
ybar,
bar width=.5cm,
width=\columnwidth,
height=.2\textwidth,
enlarge x limits=0.25,
legend style={at={(0.7,1.3)}, anchor=north,legend columns=-1},
symbolic x coords = {95,98,99},
xtick=data,
xlabel=Precision (\%),
ylabel=\texttt{Recall@P} (\%),
grid=major,
xmajorgrids=false
]
\addplot table[x=metric,y=vR]{\datatable};
\addplot table[x=metric,y=vT]{\datatable};
\addplot table[x=metric,y=vTR]{\datatable};
\addplot table[x=metric,y=vTR2]{\datatable};
\legend{\texttt{vR},\texttt{vT},\texttt{vTR},\texttt{vTR2}}
\end{axis}
\end{tikzpicture}
\caption{
\texttt{Recall@P} metrics, as per~\cite{saftescu2020kidnapped}.
}
\label{fig:ball-search}
\vspace{-.6cm}
\end{figure}

%% file: plots/db-cand.tex
\pgfplotsset{
select row/.style={
x filter/.code={\ifnum\coordindex=#1\else\def\pgfmathresult{}\fi}
}
}

\pgfplotstableread[col sep=comma]{
metric,vT,vR,vTR,vTR2
1,97.37373737373738,73.13131313131314,96.16161616161617,98.38383838383838
2,98.7878787878788,83.03030303030303,96.76767676767678,99.19191919191918
3,99.19191919191918,87.67676767676768,97.77777777777777,99.19191919191918
}\datatable

\begin{figure}
\centering
\begin{tikzpicture}
\begin{axis}[
ybar,
bar width=.5cm,
width=\columnwidth,
height=.2\textwidth,
enlarge x limits=0.25,
legend style={at={(0.7,1.3)}, anchor=north,legend columns=-1},
symbolic x coords = {1,2,3},
xtick=data,
xlabel=N - Database candidates,
ylabel=\texttt{Recall@N} (\%),
grid=major,
xmajorgrids=false
]
\addplot table[x=metric,y=vR]{\datatable};
\addplot table[x=metric,y=vT]{\datatable};
\addplot table[x=metric,y=vTR]{\datatable};
\addplot table[x=metric,y=vTR2]{\datatable};
\legend{\texttt{vR},\texttt{vT},\texttt{vTR},\texttt{vTR2}}
\end{axis}
\end{tikzpicture}
\caption{
\texttt{Recall@N} metrics, as per~\cite{barnes2020rrcd}.
}
\label{fig:db-cand}
\vspace{-.5cm}
\end{figure}

%% file: main.bbl
\begin{thebibliography}{1}
\providecommand{\url}[1]{#1}
\csname url@samestyle\endcsname
\providecommand{\newblock}{\relax}
\providecommand{\bibinfo}[2]{#2}
\providecommand{\BIBentrySTDinterwordspacing}{\spaceskip=0pt\relax}
\providecommand{\BIBentryALTinterwordstretchfactor}{4}
\providecommand{\BIBentryALTinterwordspacing}{\spaceskip=\fontdimen2\font plus
\BIBentryALTinterwordstretchfactor\fontdimen3\font minus
  \fontdimen4\font\relax}
\providecommand{\BIBforeignlanguage}[2]{{%
\expandafter\ifx\csname l@#1\endcsname\relax
\typeout{** WARNING: IEEEtran.bst: No hyphenation pattern has been}%
\typeout{** loaded for the language `#1'. Using the pattern for}%
\typeout{** the default language instead.}%
\else
\language=\csname l@#1\endcsname
\fi
#2}}
\providecommand{\BIBdecl}{\relax}
\BIBdecl

\bibitem{kim2020mulran}
G.~Kim, Y.~S. Park, Y.~Cho, J.~Jeong, and A.~Kim, ``{Mulran: Multimodal range
  dataset for urban place recognition},'' in \emph{International Conference on
  Robotics and Automation}, 2020.

\bibitem{saftescu2020kidnapped}
{\c{S}}.~S{\u{a}}ftescu, M.~Gadd, D.~De~Martini, D.~Barnes, and P.~Newman,
  ``{Kidnapped Radar: Topological Radar Localisation using
  Rotationally-Invariant Metric Learning},'' \emph{arXiv preprint
  arXiv:2001.09438}, 2020.

\bibitem{barnes2020under}
D.~Barnes and I.~Posner, ``{Under the Radar: Learning to Predict Robust
  Keypoints for Odometry Estimation and Metric Localisation in Radar},''
  \emph{arXiv preprint arXiv:2001.10789}, 2020.

\bibitem{gadd2020lookaroundyou}
M.~Gadd, D.~De~Martini, and P.~Newman, ``{Look Around You: Sequence-based Radar
  Place Recognition with Learned Rotational Invariance},'' \emph{arXiv preprint
  arXiv:2003.04699}, 2020.

\bibitem{demartini2020kradar}
D.~De~Martini, M.~Gadd, and P.~Newman, ``{kRadar++: Coarse-to-fine FMCW
  Scanning Radar Localisation},'' \emph{Sensors}, vol.~20, no.~21, p. 6002,
  2020.

\bibitem{wang2021radarloc}
W.~Wang, P.~P. de~Gusmo, B.~Yang, A.~Markham, and N.~Trigoni, ``{RadarLoc:
  Learning to Relocalize in FMCW Radar},'' \emph{arXiv preprint
  arXiv:2103.11562}, 2021.

\bibitem{ye2019unsupervised}
M.~Ye, X.~Zhang, P.~C. Yuen, and S.-F. Chang, ``{Unsupervised Embedding
  Learning via Invariant and Spreading Instance Feature},'' \emph{arXiv
  preprint arXiv:1904.03436}, 2019.

\bibitem{simonyan2014very}
K.~Simonyan and A.~Zisserman, ``{Very deep convolutional networks for
  large-scale image recognition},'' \emph{arXiv preprint arXiv:1409.1556},
  2014.

\bibitem{barnes2020rrcd}
D.~Barnes, M.~Gadd, P.~Murcutt, P.~Newman, and I.~Posner, ``{The Oxford Radar
  RobotCar Dataset: A Radar Extension to the Oxford RobotCar Dataset},''
  \emph{arXiv preprint arXiv:1909.01300}, 2019.

\end{thebibliography}
